\documentclass[10pt, a4paper]{article}
\usepackage{lrec2022} 
\usepackage{multibib}
\newcites{languageresource}{Language Resources}
\usepackage{graphicx}
\usepackage{balance}
\usepackage{tabularx}
\usepackage{soul}
\usepackage{titlesec}
\titleformat{\section}{\normalfont\large\bfseries\center}{\thesection.}{1em}{}
\titleformat{\subsection}{\normalfont\SmallTitleFont\bfseries\raggedright}{\thesubsection.}{1em}{}
\titleformat{\subsubsection}{\normalfont\normalsize\bfseries\raggedright}{\thesubsubsection.}{1em}{}
\renewcommand\thesection{\arabic{section}}
\renewcommand\thesubsection{\thesection.\arabic{subsection}}
\renewcommand\thesubsubsection{\thesubsection.\arabic{subsubsection}}

\usepackage{epstopdf}
\usepackage[utf8]{inputenc}

\usepackage{hyperref}
\usepackage{xstring}

\usepackage{color}
\usepackage{subcaption}
\usepackage{graphicx}
\usepackage{booktabs}
\usepackage{amsmath}
\usepackage{enumitem}
\usepackage{xspace}

\usepackage{arydshln}

\newcommand{\moveup}{\vspace*{-2mm}}
\newcommand{\moveups}{\vspace*{-1mm}}
\newcommand{\tabcaption}[1]{\vspace*{-1mm}\caption{#1}\vspace*{-3mm}}
\newcommand{\figcaption}[1]{\vspace*{-3mm}\caption{#1}\vspace*{-4mm}}
\newcommand{\ignore}[1]{}
\newcommand{\xhdr}[1]{\vspace{1.3mm}\noindent{{\bf #1.}}}

\newcommand{\ie}{\textit{i.e.}\xspace}
\newcommand{\eg}{\textit{e.g.}\xspace}

\title{Efficient Entity Candidate Generation for Low-Resource Languages}

\name{Alberto Garc\'{i}a-Dur\'{a}n,
  Akhil Arora,
  Robert West} 

\address{
        EPFL, Switzerland \\
        agaduran@gmail.com \\
        \{akhil.arora, robert.west\}@epfl.ch}

\abstract{
Candidate generation is a crucial module in entity linking. It also plays a key role in multiple NLP tasks that have been proven to beneficially leverage knowledge bases. Nevertheless, it has often been overlooked in the monolingual English entity linking literature, as na\"ive approaches obtain very good performance. Unfortunately, the existing approaches for English cannot be successfully transferred to poorly resourced languages.
This paper constitutes an in-depth analysis of the candidate generation problem in the context of cross-lingual entity linking with a focus on low-resource languages. Among other contributions, we point out limitations in the evaluation conducted in previous works. We introduce a characterization of queries into types based on their difficulty, which improves the interpretability of the performance of different methods.
We also propose a \emph{light-weight} and \emph{simple} solution based on the construction of indexes whose design is motivated by more complex transfer learning based neural approaches. A thorough empirical analysis on 9 real-world datasets under 2 evaluation settings shows that our simple solution outperforms the state-of-the-art approach in terms of both quality and efficiency for almost all datasets and query types.
 \\ \newline  }

\begin{document}

\maketitleabstract

\section{Introduction}
\label{sec:intro}

Candidate generation \cite{shen2014entity,Zhou2020ImprovingCG} is the task of retrieving a short list of plausible candidate entities from a knowledge base (KB) for a given mention. This module is usually a fundamental part of a more elaborate pipeline as that of entity linking \cite{guo2013improving}, coreference resolution \cite{Zhang2019KnowledgeawarePC} or question answering \cite{Bao2016ConstraintBasedQA}. With few exceptions \cite{Gillick2019LearningDR}, it is present in almost all existing entity linking techniques, where the subsequent entity disambiguation technique only has to evaluate a small set of entities from the KB, instead of all of them. Therefore, in addition to obtaining a high recall---fraction of mentions where the true entity is present in the set of candidates---a desired characteristic of a candidate generator is to constitute a low computational overhead of the full pipeline. Cross-lingual entity linking \cite{Tsai2016CrosslingualWU,Sil2018NeuralCE} associates mentions in non-English documents to entities from a KB---typically the English Wikipedia. This poses several challenges that can often be attributed to the amount of resources of the target language, especially if it has low resources. Candidate generation is deemed trivial in the English entity linking literature, as there exist resources (\ie indexes/lookup tables \cite{Spitkovsky2012ACD}) capable of providing high candidate recall. However, this is not the case in the cross-lingual setting. This is specially true when dealing with low-resource languages (\eg , karelian or turkmen), wherein, moreover, the number of works that tackle this important step is very scarce. Some recent approaches \cite{Rijhwani2019ZeroshotNT,Zhou2020ImprovingCG} are based on the idea of performing transfer learning from a pivot language that is lexically similar to the target language.

\xhdr{Contributions} In this paper we propose a \textit{light-weight} solution, called pivot token indexes (\textsc{pti}), to the candidate generation problem with a special focus on low-resource languages. \textsc{pti} combines prior and posterior probabilities learned by aggregating information from the pivot and target language. These probabilities are constructed with tokens the mentions can be decomposed into. 
This simple solution keeps the efficiency of standard indexes---thereby possessing the desired property of a candidate generator---while being equipped with the sophistication of more complex solutions \cite{Zhou2020ImprovingCG}. 
Most importantly, the main contribution of this work is an in-depth analysis of the candidate generation problem in the context of entity linking, to which we make several contributions: i) we establish that the (inherited) practice of ignoring mentions that are not linkable to the English Wikipedia leads to an unrealistic evaluation; ii) we perform a realistic evaluation and propose an efficient trick to increase the ceiling recall of \textit{any} candidate generator; iii) we introduce three query types according to their difficulty based on the available training data; iv) we compare the performance of the different methods and establish differences across them based on the amount of resources of the pivot language and the query type. Moreover, our experimental section includes experiments on 9 datasets and two different learning settings (without and with some supervision from the target language). We also empirically analyze the complexity of \textsc{pti} and conclude that it takes up much less memory, and requires significantly less training and inference time than the current state-of-the-art \cite{Zhou2020ImprovingCG}.

\xhdr{Other applications of candidate generators} In general, tasks that can beneficially leverage the knowledge contained in KBs may benefit from better candidate generators. Examples of such tasks are coreference resolution \cite{Zhang2019KnowledgeawarePC} or question answering \cite{Bao2016ConstraintBasedQA}, to name but a few. These works retrieve facts from a KB that are related to a given mention. As in the case of entity linking, candidate generation may alleviate the computational cost of the subsequent modules by just keeping a reduced set of facts. 

\section{Background}
\label{sec:background}
Let $m$ be a mention observed in a document, and let $e$ be the ground truth entity associated to that mention. The task of a candidate generator $\mathcal{CG}^K$ is to retrieve a list of $K$ possible candidate entities such that $e \in \mathcal{CG}^K(m)$. The candidate generator is trained with mention entity pairs observed in a training corpus $\mathcal{C}_{\textsc{lang}}$, where the mentions are written in a certain language $\textsc{lang}$. At test time, candidate generators provide a score to all the entities that form the candidate space, and find the $K$ entities with the largest scores---this is a linear operation (done with a max-heap structure) with respect to the candidate space.

A candidate generator must have good recall, and also be much more computational efficient as compared to the more complex downstream model. See \cite{ling2015design} for a detailed discussion. We use \textsc{en}, \textsc{pl} and \textsc{tl} to indicate English, pivot and target language, respectively.

\section{Related Work}
\label{sec:rel_work}
While candidate generation in monolingual (English) entity linking is deemed a trivial task given the existence of high-quality mention-entity lookup tables, most of other languages---specially low-resource languages---do not benefit from lookup tables of a similar quality. Therefore, contrary to (English) entity linking, candidate generation constitutes a challenge in the cross-lingual entity linking literature. 



\subsection{Wikipedia-based Indexes}
This is the \textit{de facto} standard for candidate generation in the entity linking literature \cite{Sil2016OneFA,Yamada2017LearningDR,Sil2018NeuralCE,Upadhyay2018JointMS}. An index of prior probabilities $P(\text{entity}|\text{mention})$ is constructed from a corpus. While this solution has been shown to provide a good recall for languages where the training corpus is large, for low-resource languages the training corpus may offer a very poor coverage in terms of both mentions \cite{Fu2020DesignCF} and entities.  

\subsection{Pivoting-based Embedding Approaches}
\label{sec:charagram}
Methods within this category leverage information of a pivot language that is lexically related to the target language: the candidate generator is trained with data from the pivot language, but then it is evaluated in the target language. These methods rely on a certain lexical overlap between pairs of related languages (\eg Italian and Corsican) that allow to transfer learning between the pivot---which is supposed to have more resources than the target language---and the target language. This is opposed to other techniques \cite{gollins2001improving} that rely on the existence of accurate machine translation models to map all the resources to a common language. Unfortunately, the performance of translation models is also largely affected by the scarce data that is available for low-resource languages. Similar ideas have been applied to other problems such as POS tagging \cite{Fang2017ModelTF} or machine translation \cite{Zoph2016TransferLF}. 

The learning objective consists of maximizing a similarity function $sim$ (the cosine similarity) between mention $\mathbf{m}$ and entity $\mathbf{e}$ embedding pairs:
\begin{align}
\moveup
\moveups
    \mathcal{L}_{\textsc{pl}} = \sum_{(m, e) \in \mathcal{C}_{\textsc{pl}}} &\max{(0, 1 - sim(\mathbf{m}, \mathbf{e}^{\textsc{en}^+})} \nonumber \\
    & \qquad + sim(\mathbf{m}, \mathbf{e}^{\textsc{en}^-}) ),
    \label{eq:charagram}
    \moveup
\end{align}
where $\mathcal{C}_{\textsc{pl}}$ is the collection of training mention entity pairs in the pivot language, and $\mathbf{m} = \texttt{Enc}(\texttt{tknr}(m))$ is the mention embedding obtained by applying an encoding function \texttt{Enc} to the output of a tokenizer $\texttt{tknr}(m)$. A similar architecture is used to obtain the entity embedding from its English name. The superscript $+$ or $-$ indicates whether the entity is the target or a randomly sampled one. 

Previous works have explored different tokenizers (\eg character-level \cite{Rijhwani2019ZeroshotNT} or character $n$-grams \cite{Zhou2020ImprovingCG}) and encoding functions (\eg LSTM- \cite{Rijhwani2019ZeroshotNT} or CNN-based architectures \cite{Zhang2015CharacterlevelCN}). The best setup \cite{Zhou2020ImprovingCG} consists of a character $n$-gram tokenizer, followed by an encoder that maps each token to an embedding (via an embedding lookup table), sums them all up, and applies a nonlinear activation function to the aggregated representation. This approach, called \textsc{Charagram} \cite{Wieting2016CharagramEW}, has shown to also outperform other type of approaches based on indexes and translation  \cite{Pan2017CrosslingualNT}.\\

\noindent\textbf{Additional Related Work.} There are related tasks such as transliteration \cite{Upadhyay2018BootstrappingTW} and bilingual lexical induction \cite{Haghighi2008LearningBL}. The former is a harder task, as it consists of generating, instead of retrieving, English entity names. The latter relies to a great extent on the existence of large monolingual corpus in both languages \cite{Vulic2020AreAG}. 

\section{\textsc{Pivot Token Indexes}}
\label{sec:elt}
Intuitively, \textsc{Charagram} learns to associate tokens that co-occur between mentions and English entity names. The learned associations between tokens determine the similarity (score) between entity and mention embeddings. This approach has been shown to provide good recall but both memory requirements and training/inference time are considerable (an empirical analysis is in Section \ref{sec:analysis}). Thus, it is arguable whether \textsc{Charagram} fulfills a desired characteristic of a candidate generator, namely its computational overhead is low. Similarly to \textsc{Charagram}, our approach also learns associations between mentions and entities. Different to \textsc{Charagram}, the associations learned by \textsc{pti} are meant to indicate the attachment strength between tokens and entities, and not equivalences between tokens. \textsc{pti} consists of two indexes---then being high efficient---that are linearly combined to score entities. The indexes correspond to the prior probability that associates tokens to entities, and the posterior probability that associates entities to tokens. The goal of the posterior probability is to counterbalance the (low) score that unpopular entities obtain for popular tokens from the prior index. \textsc{pti} has the following steps: 
\begin{enumerate}
    \item We apply a tokenizer \texttt{tknr} to the collection of mentions observed in the training corpus of the pivot language $\mathcal{C}_{\textsc{pl}}$. Each mention $m$, linked to a certain entity, is decomposed into a set of $n_m$ tokens $\texttt{tknr}(m) = \{\text{token}_1,\text{token}_2, \cdots,\text{token}_{n_m}\}$.
    \item We construct prior $P(\text{entity}|\text{token})$ and posterior $P(\text{token}|\text{entity})$ token-based probabilities. Optionally, we can remove low probabilities by thresholding (Section \ref{sec:analysis}) the indexes without an impact in the performance. The thresholding significantly reduces the memory requirements of \textsc{pti}.
    \item At test time, given a mention $m$, each entity $ent$ obtains an score as follows:
    \begin{align}
        &\sum_{\text{tok} \in \texttt{tknr}(m)} P(ent|\text{tok}) + \lambda  P(\text{tok}|ent) \nonumber \\ 
        & = \sum_{\text{tok} \in \texttt{tknr}(m)} P(ent|\text{tok}) (1 + \lambda  \cfrac{P(\text{tok})}{P(ent)}),
    \label{eq:scoring}
    \end{align}
    where $\lambda$ is a weighting term. As argued, the posterior probability term may counterbalance the bias introduced by the prior probability towards the popular entities: a large value of $\lambda$ will favor unpopular entities. While these scores do not have a probabilistic interpretation, one can easily normalize them so that they sum up to 1. As opposed to \textsc{Charagram}, \textsc{pti} performs a search that only includes the entities that have a non-zero score; Then contributing towards a better efficiency.
\end{enumerate}

We will see in Section \ref{sec:results} that this simple approach has the following advantages as compared to \textsc{Charagram}: fewer parameters, less training time, less inference time, and better performance in almost all settings. 
 
 \subsection{Joint learning with the target language}
In practice, there is often some training data that is available for the target language. To incorporate this data into the construction of the prior and posterior probabilities we modify step 2 by weighting the counts over the pivot language corpus with a factor $\alpha$. Thus, the index of priors is computed as
\begin{align*}
P&(\text{entity}|\text{token}) = \\ \nonumber
&=\cfrac{\#^{\textsc{tl}}\text{(entity}|\text{token)} + \alpha \#^{\textsc{pl}}\text{(entity}|\text{token)}}{\sum_{e \in \mathcal{E}}(\#^{\textsc{tl}}\text{(e}|\text{token)} + \alpha \#^{\textsc{pl}}\text{(e}|\text{token)})}, \nonumber
\end{align*}
where $\#^{\textsc{lang}}(\cdot)$ is a shortcut that refers to the number of times the conditional argument is observed in the corpus $\mathcal{C}_{\textsc{lang}}$. $\mathcal{E}$ corresponds to the set of entities observed in the corpora. The same weighting is applied to compute the posterior probabilities. At test time, entities are scored following Eq. (\ref{eq:scoring}).

More complex versions of our approach \textsc{pti} are in Appendix \ref{app:failed_ext}. While these versions led to minor improvements, their complexity also increased.

\section{Experimental Setup}
\label{sec:exps}
All the resources required to reproduce the experiments are available at~\url{https://github.com/epfl-dlab/pti-candgen}.

In order to comprehensively understand the performance of \textsc{pti} we run experiments on a variety of publicly available datasets. The benchmarked methods are compared in terms of accuracy and complexity (number of parameters and training/inference time). The accuracy is measured by the recall, which is defined as the proportion of retrieved candidate lists containing the correct entity. More formally, the recall@$K$ is defined as 
\begin{equation}
    \moveups
    \text{recall}@K = \cfrac{\sum_{(m,e) \in \mathcal{T}} \delta(e \in \mathcal{CG}^K (m))}{|\mathcal{T}|}*100 \nonumber
    \moveups
\end{equation}
where $\mathcal{T}$ is the set of mentions entity pairs in an evaluation set, and $\delta(\cdot)$ returns 1 if its argument is true, and 0 otherwise. As in previous work \cite{Zhou2020ImprovingCG} we set $K=30$ for our main set of experiments. However, for a subset of the experiments we include a comparative performance for other values of $K$ in Appendix \ref{app:recall}.

We compare the performance of the different methods in the two following learning settings:

\begin{itemize}[nolistsep,leftmargin=4mm]
    \item \textbf{Zero-shot}. We use zero resources---no training data---from the target language. Therefore, the candidate generator must fully rely on the available information for the pivot language. While it is possible to obtain a (small) training set of mention entity pairs for some target languages, in some other cases the target language is so poorly resourced that this is the only possible learning setting.
    \item \textbf{Joint learning with the target language}. Even for low-resource languages, there is often some training data for the target language that can be used in conjunction with the data that is available for the pivot language. We sometimes refer to this setting as supervised.
\end{itemize}

\noindent We note that the aim of this section is to establish a detailed comparison of different candidate generation methods. As any progress in the candidate generator is independent from any improvement in the entity disambiguation module, the integration of the candidate generation in the entity linking pipeline is out of the scope of this work.

\begin{figure}[t]
\includegraphics[width=\columnwidth]{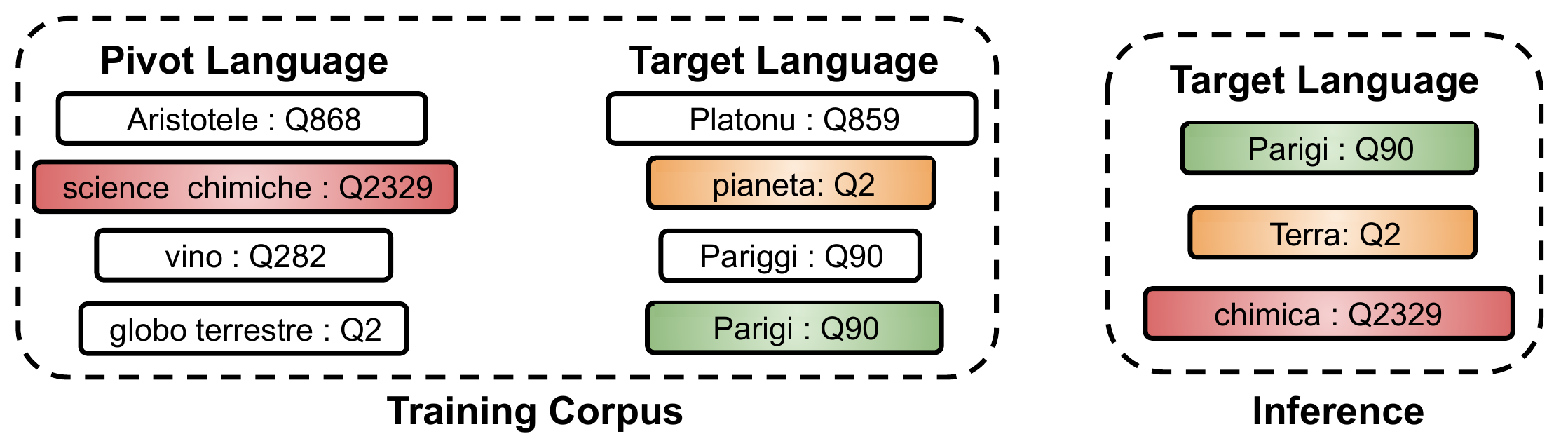}
\caption{\label{fig:queries_type} Query types according to their difficulty. The target and pivot language correspond to corsican (\textsc{co}) and italian (\textsc{it}), respectively. The backgrounds of the queries (\textit{Mention : Ground Truth Entity}) indicate their type (\textsc{Easy}=green, \textsc{Medium}=orange, and \textsc{Hard}=red).}
\end{figure}

\subsection{Query Types}
\label{sec:query}
As discussed in Section \ref{sec:exps}, despite the little resources that are available for the target languages, sometimes it is possible to obtain some training data. This training data will determine the difficulty of the queries contained in the test set. Given a mention query $m$ that links to an entity $e$, we classify it according to our proposed categorization (see Figure \ref{fig:queries_type} for a visual explanation):
\begin{itemize}[nolistsep,leftmargin=4mm]
    \item \textsc{Easy}. The mention $m$ is observed with the entity $e$ in the training data of the target language.
    \item \textsc{Medium}. The entity $e$ is observed in the training data of the target language but with mentions other than $m$.
    \item \textsc{Hard}. The entity $e$ is never observed in the training data of the target language. To successfully link these mentions we fully rely on the pivot language. \textit{In the zero-shot setting, all mention queries belong to this type}.
\end{itemize}

\subsection{Baselines}
\label{sec:baselines}


\begin{itemize}[nolistsep,leftmargin=4mm]
\item \textsc{WikiPriors}. Prior probabilities (\ie $P(\text{entity}|\text{mention})$) are estimated using a training corpus obtained from both a target and a pivot language, and constructs indexes mapping mentions to their co-occurring entities. This simple procedure resembles the approach followed by \cite{Spitkovsky2012ACD}, where a lookup table was built for entities that have an English Wikipedia page. The strength of a mention-entity association is given by the estimated prior probabilities. In the supervised setting, it first generates candidates using the index constructed on the target language. More details are provided in Appendix \ref{app:wikipriors}.


\item\textsc{Charagram} \cite{Zhou2020ImprovingCG}. This is the state-of-the-art on candidate generation for (low-resource) cross-lingual settings. It has shown to significantly outperform other architectures such as \cite{Rijhwani2019ZeroshotNT}, and translation-based methods \cite{Pan2017CrosslingualNT}, among others. For the supervised setting, we extend \textsc{Charagram} to also incorporate training data from the target language by optimizing the loss $\mathcal{L}_{\textsc{tl}} + \mu \mathcal{L}_{\textsc{pl}}$. More details are provided in Section \ref{sec:charagram}.
\end{itemize}

\subsection{Datasets}
\label{sec:data}
As previous works \cite{Rijhwani2019ZeroshotNT,Zhou2020ImprovingCG}, we create our datasets from Wikipedia. We choose Wikipedia as, contrary to other datasets (\eg DARPA LORELEI \cite{Strassel2016LORELEILP}), it can be accessed without any restriction, and, importantly, its crowdsourcing spirit guarantees a realistic setting.

We select 9 pairs of languages, where the target language is always (extremely) low-resourced, and the pivot language can be categorized as either high, medium, or low-resourced. The distinction regarding the amount of resources of the pivot language has never done in previous works, but it helps to analyze the performance of the different methods. The 9 language pairs (\textsc{target}-\textsc{pivot}) are Corsican-Italian (\textsc{co}-\textsc{it}), Limburgish-Dutch (\textsc{li}-\textsc{nl}), Bavarian-German (\textsc{bar}-\textsc{de}), Karelian-Finnish (\textsc{olo}-\textsc{fi}), Macedonian-Bulgarian (\textsc{mk}-\textsc{bg}), Javanese-Indonesian (\textsc{jv}-\textsc{id}), Turkmen-Azerbaijani (\textsc{tk}-\textsc{az}), Marathi-Hindi (\textsc{mr}-\textsc{hi}) and Faroese-Icelandic (\textsc{fo}-\textsc{is}). Appendix \ref{app:dataset} contains information about the dataset statistics (Table \ref{table:data}), additional preprocessing details and detailed information about the selection of the pivot languages.

\begin{table*}[h!]
\centering
\normalsize
\vspace{0.1cm}
\tabcaption{\label{table:main_results} Micro recall@30 of different approaches. The upper and lower block show the performance of the methods in the zero-shot and supervised setting, respectively. For the pairs \textsc{co - it}, \textsc{olo - fi} and \textsc{tk - az}, where there is no training data in the target language, the supervised setting amounts to the zero-shot one. The best performance for each target language and setting is in \textbf{bold}.}
\resizebox{0.9\linewidth}{!}{
    \begin{tabular}{lrrrrrrrrr}
    \toprule
            & \multicolumn{9}{c}{\textbf{Pivot Language}} \\ \cmidrule(lr){2-10}
          &  \multicolumn{3}{c}{High Resource}     & \multicolumn{3}{c}{Medium Resource}         & \multicolumn{3}{c}{Low Resource} \\ \cmidrule(lr){2-4} \cmidrule(lr){5-7} \cmidrule(lr){8-10}
     \textbf{Target Lang.} &  \textsc{co} - \textsc{it}   & \textsc{li} - \textsc{nl}  &    \textsc{bar} - \textsc{de} &  \textsc{olo} - \textsc{fi}   & \textsc{mk} - \textsc{bg}  &    \textsc{jv} - \textsc{id} &  \textsc{tk} - \textsc{az}   & \textsc{mr} - \textsc{hi}  &    \textsc{fo} - \textsc{is}  \\ \midrule \midrule
     \multicolumn{10}{c}{\textsc{Zero-Shot}} \\  \midrule
     \textsc{WikiPriors} & 47 & 51 & 63 & 45 & 31 & 74 & 48 & 30 & 29 \\
     \textsc{Charagram} & 51 & 51 & 50 & 48 & \textbf{47} & 60 & \textbf{67} & \textbf{48} & \textbf{49}\\ \cmidrule(lr){1-10}
     \textsc{pti} & \textbf{63} & \textbf{70} & \textbf{76} & \textbf{72} & \textbf{49} & \textbf{79} & \textbf{69} & 40 & \textbf{47} \\
        \bottomrule \midrule
     \multicolumn{10}{c}{\textsc{Joint learning with the target language}} \\  \midrule
     \textsc{WikiPriors} & 47 & 68 & 77 & 45 & 59 & 82 & 48 & 58 & 58 \\
     \textsc{Charagram} &  51 & 72 & 73 & 48 & \textbf{69} & 72 & \textbf{67} & 66 & 74\\ \cmidrule(lr){1-10}
     \textsc{pti} & \textbf{63} & \textbf{84} & \textbf{88} & \textbf{72} & \textbf{71} & \textbf{87} & \textbf{69} & \textbf{70} & \textbf{79} \\
        \bottomrule
    \end{tabular}
}
\end{table*}

\begin{table*}[t]
\centering
\small
\tabcaption{\label{table:results_types} Performance breakdown in terms of recall@30 for the case when the models are trained with some supervision from the target language---except for \textsc{co - it}, \textsc{olo - fi} and \textsc{tk - az}, which correspond to the zero-shot setting. \textsc{E}, \textsc{M} and \textsc{H} are shortcuts that refer to \textsc{Easy}, \textsc{Medium} and \textsc{Hard} queries, respectively. The best performance for each target language and query type is shown in \textbf{bold}.}
\resizebox{0.99\linewidth}{!}{
    \begin{tabular}{lcrrrrrrcrrrrrrcrrrrrr}
    \toprule
            & \multicolumn{21}{c}{\textbf{Pivot Language}} \\ \cmidrule(lr){2-22}
          &  \multicolumn{7}{c}{High Resource}     & \multicolumn{7}{c}{Medium Resource}         & \multicolumn{7}{c}{Low Resource} \\ \cmidrule(lr){2-8} \cmidrule(lr){9-15} \cmidrule(lr){16-22}
     \textbf{Target Lang.} & \multicolumn{1}{c}{\textsc{co} - \textsc{it}} & \multicolumn{3}{c}{\textsc{li} - \textsc{nl}}  &    \multicolumn{3}{c}{\textsc{bar} - \textsc{de}} & \multicolumn{1}{c}{\textsc{olo} - \textsc{fi}} & \multicolumn{3}{c}{\textsc{mk} - \textsc{bg}}  &    \multicolumn{3}{c}{\textsc{jv} - \textsc{id}}  & \multicolumn{1}{c}{\textsc{tk} - \textsc{az}} & \multicolumn{3}{c}{\textsc{mr} - \textsc{hi}}  &  \multicolumn{3}{c}{\textsc{fo} - \textsc{is}}  \\ 
     \cmidrule(lr){2-2} \cmidrule(lr){3-5} \cmidrule(lr){6-8} \cmidrule(lr){9-9} \cmidrule(lr){10-12} \cmidrule(lr){13-15} \cmidrule(lr){16-16} \cmidrule(lr){17-19} \cmidrule(lr){20-22} 
     \textbf{Query Type} & \textsc{H} & \textsc{E} & \textsc{M} & \textsc{H} & \textsc{E} & \textsc{M} & \textsc{H} & \textsc{H} & \textsc{E} & \textsc{M} & \textsc{H} & \textsc{E} & \textsc{M} & \textsc{H} & \textsc{H} & \textsc{E} & \textsc{M} & \textsc{H} & \textsc{E} & \textsc{M} & \textsc{H} \\ 
          \midrule    
    \textsc{WikiPriors} & 47 & \textbf{100} & 49 & 57 & \textbf{100} & 56 & 74 & 45 & \textbf{99} & 53 & 26 & \textbf{100} & 79 & 68 & 48 & \textbf{99} & 61 & 15 & \textbf{100} & 38 & 35 \\
     \textsc{Charagram} & 51 & 83 & 72 & 61 & 84 & 75 & 62 & 48 & 86 & 74 & \textbf{46} & 82 & 70 & 61 & \textbf{67} & 82 & 75 & \textbf{42} & 90 & 80 & \textbf{54}\\ \cmidrule(lr){1-22}
     \textsc{pti} & \textbf{63} & \textbf{98} & \textbf{80} & \textbf{73} & \textbf{99} & \textbf{84} & \textbf{82} & \textbf{72} & 92 & \textbf{78} & \textbf{44} & \textbf{99} & \textbf{86} & \textbf{78} & \textbf{69} & \textbf{98} & \textbf{82} & 30 & \textbf{100} & \textbf{93} & 45 \\
        \bottomrule
    \end{tabular}
    }
    \moveup
\end{table*}

Except for cases when zero-shot is the only possible learning setting (\textsc{co}-\textsc{it}, \textsc{olo}-\textsc{fi} and \textsc{tk}-\textsc{az}) we create the validation and test sets by sampling a maximum of 1,000 unique queries of each query type---in total it amounts to a maximum of 3,000 queries; and the training data is adapted accordingly. The same validation and test sets are used in the zero-shot setting.
For target languages that can only be evaluated in the zero-shot setting, we equally distribute the available mention entity pairs into validation and test. These sets never contain duplicates.

\subsection{Towards a Realistic Evaluation}
\label{sec:realistic}
A common practice in the (English) entity linking literature \cite{Yamada2017LearningDR,Ganea2017DeepJE} consists of ignoring mentions whose correct entity does not exist in the considered KB---typically the English Wikipedia which contains over 5 million entities. This is not a major problem in these models, as this has been shown \cite{Liu2019AttentionBasedJE} to remove a small percentage (typically less than 5\%) of the data.
Previous works \cite{Rijhwani2019ZeroshotNT,Zhou2020ImprovingCG} on candidate generation for cross-lingual entity linking have adopted the same practice. Indeed, cross-lingual entity linking is defined as the task of grounding mentions in non-English documents to entries in the English Wikipedia \cite{Tsai2016CrosslingualWU,Upadhyay2018JointMS}. However, we will show in Section \ref{sec:analysis} that this practice is problematic in the cross-lingual setting, as we would have to ignore a very large percentage of data.

\begin{figure*}[t]
     \centering
     \moveups
    \begin{subfigure}[t]{0.27\textwidth}
        \raisebox{-\height}{\includegraphics[width=\textwidth]{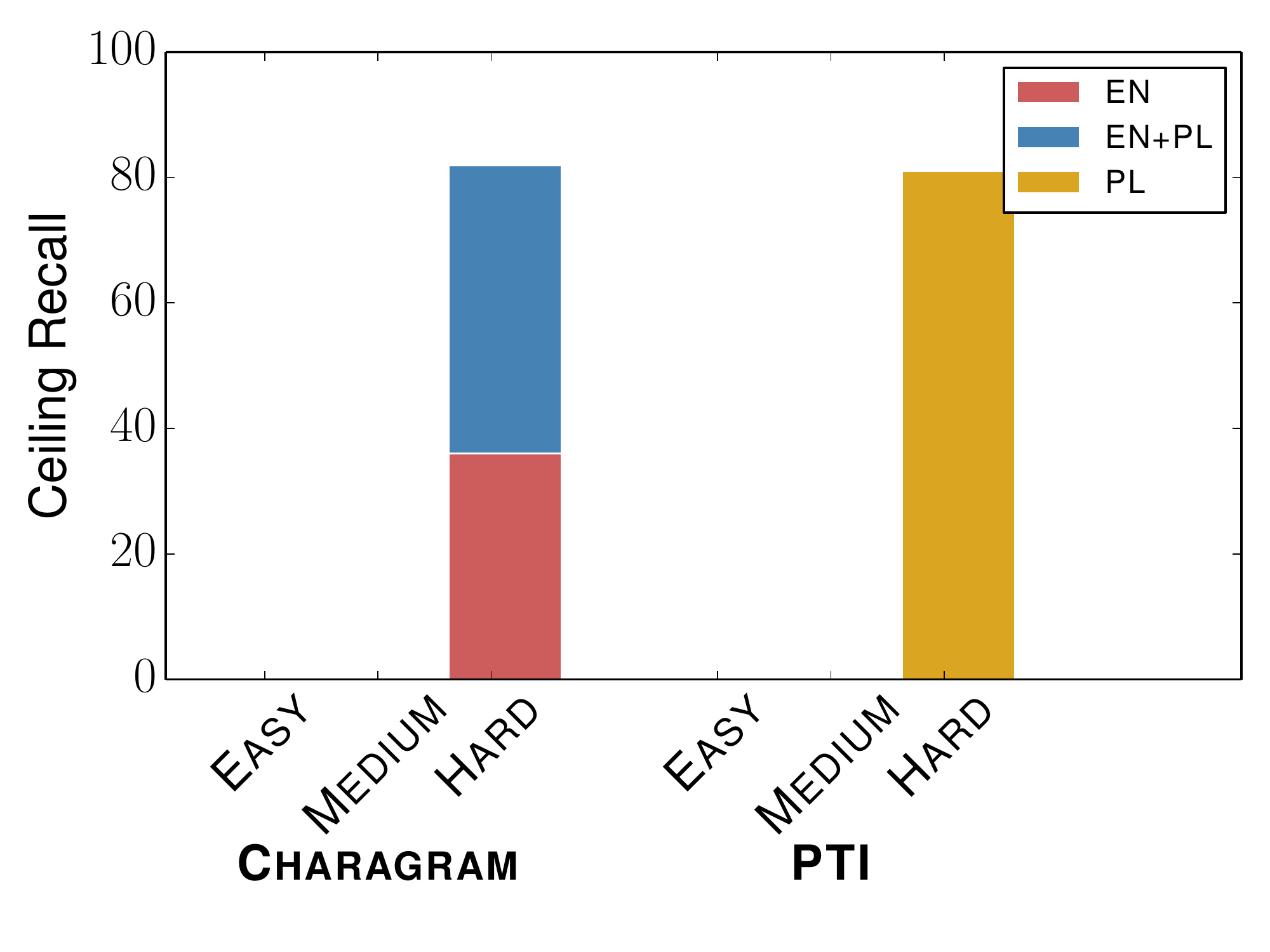}}
        \vspace{-0.4cm}\caption{\textsc{co - it}}\vspace{-0.1cm}
    \end{subfigure}
    \hfill
    \begin{subfigure}[t]{0.27\textwidth}
        \raisebox{-\height}{\includegraphics[width=\textwidth]{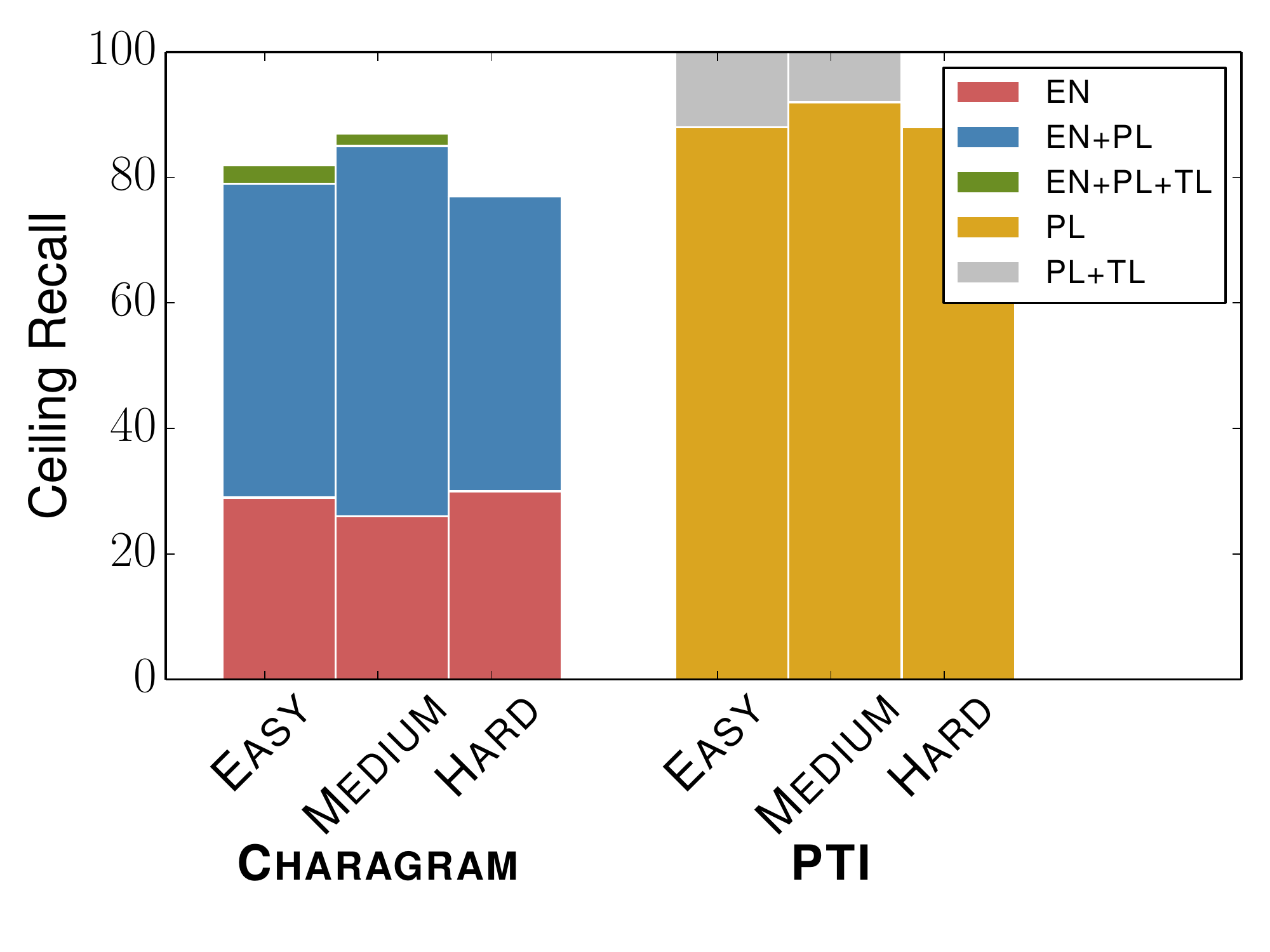}}
        \vspace{-0.4cm}\caption{\textsc{li - nl}}\vspace{-0.1cm}
    \end{subfigure}
    \hfill
    \begin{subfigure}[t]{0.27\textwidth}
        \raisebox{-\height}{\includegraphics[width=\textwidth]{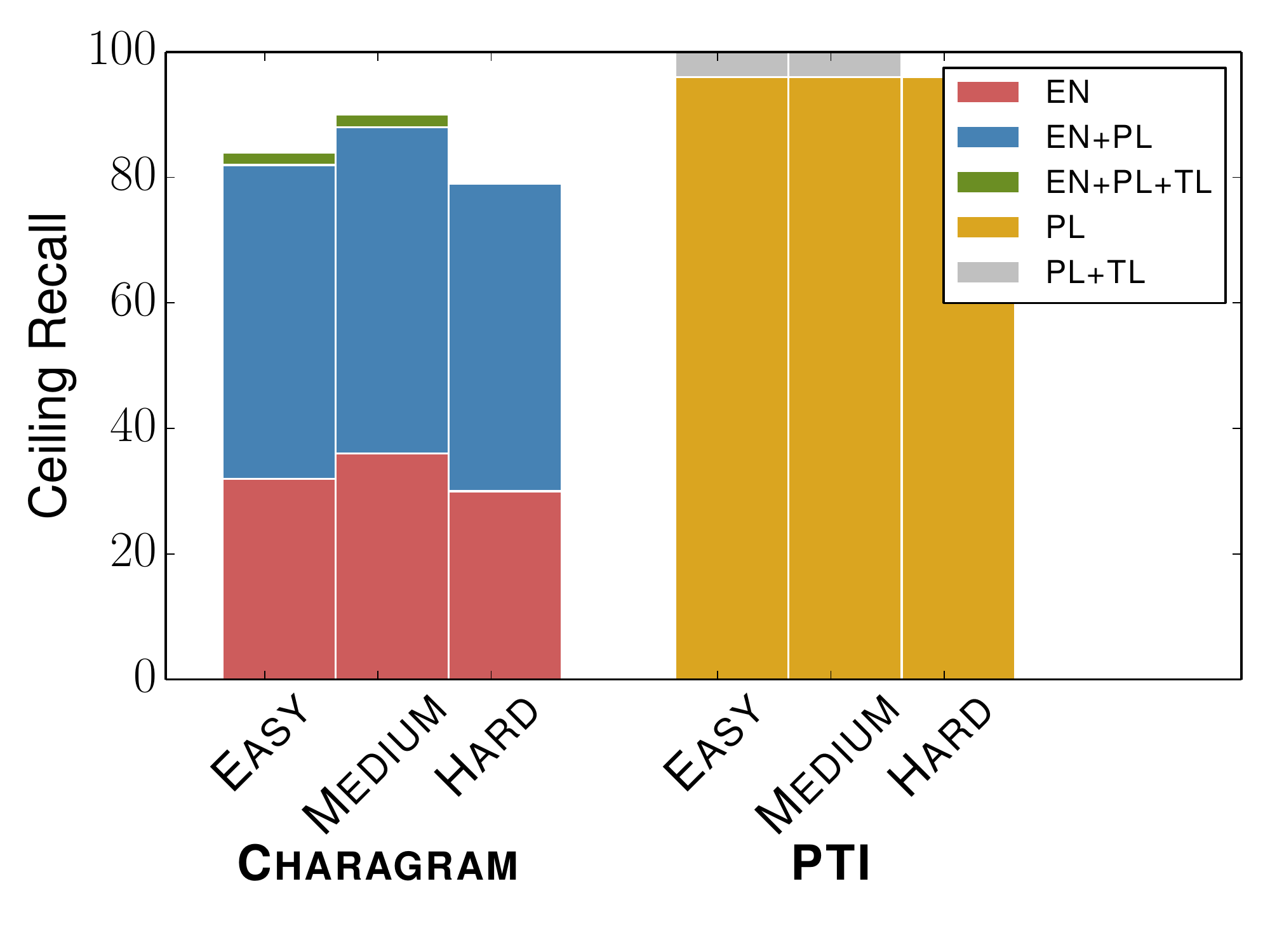}}
        \vspace{-0.4cm}\caption{\textsc{bar - de}}\vspace{-0.1cm}
    \end{subfigure}
    \begin{subfigure}[t]{0.27\textwidth}
        \raisebox{-\height}{\includegraphics[width=\textwidth]{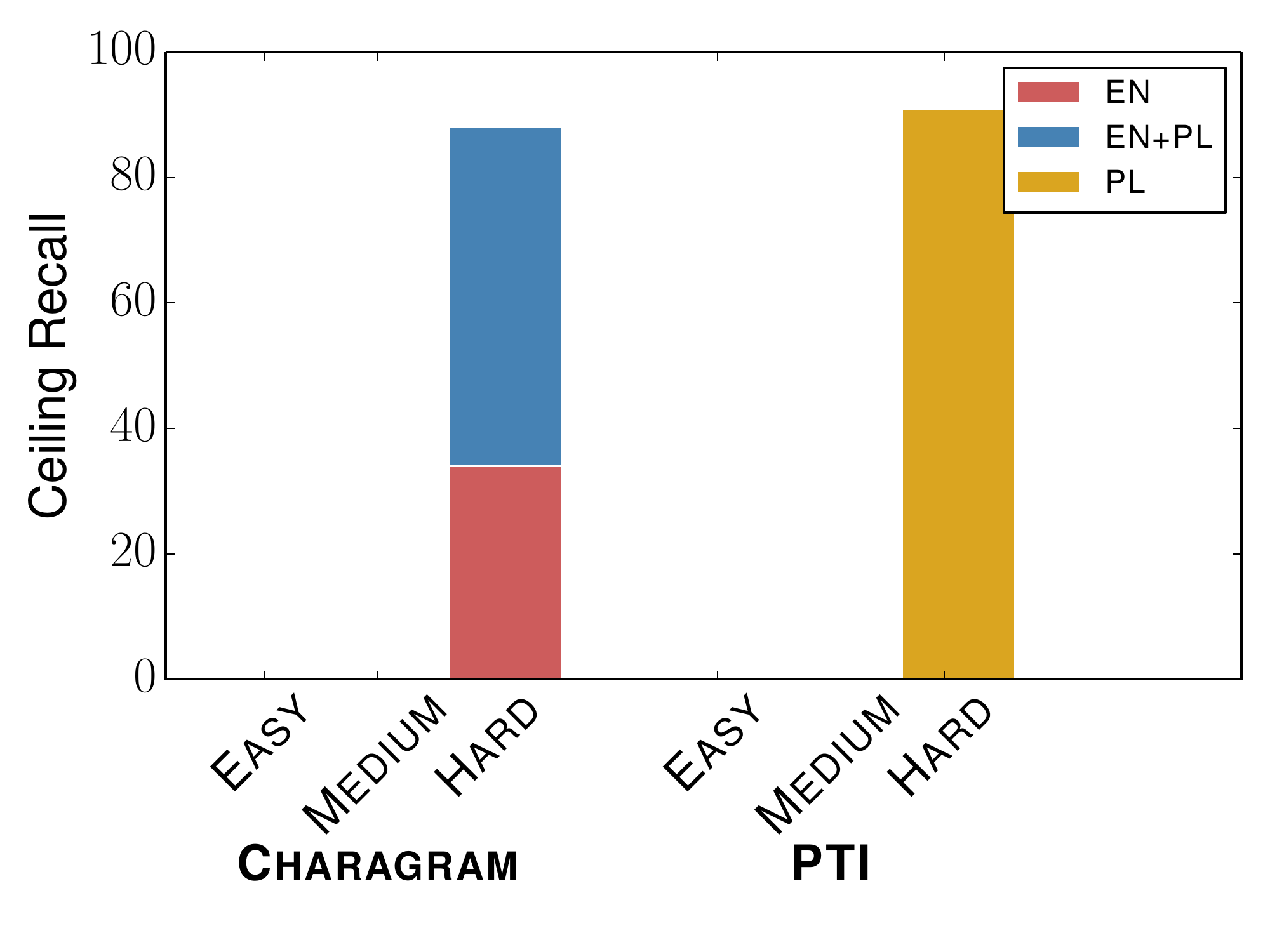}}
        \vspace{-0.4cm}\caption{\textsc{olo - fi}}\vspace{-0.1cm}
    \end{subfigure}
    \hfill
    \begin{subfigure}[t]{0.27\textwidth}
        \raisebox{-\height}{\includegraphics[width=\textwidth]{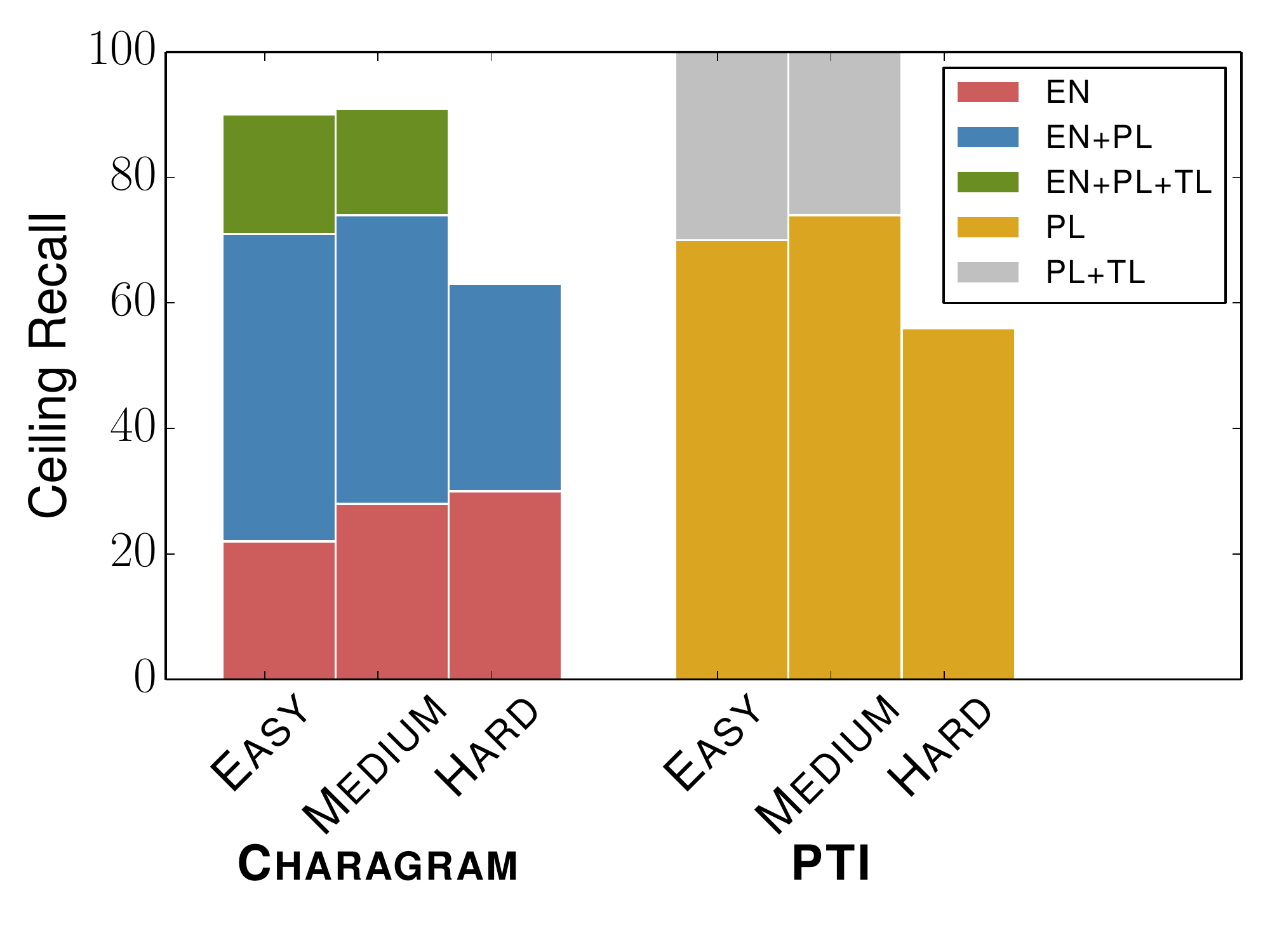}}
        \vspace{-0.4cm}\caption{\textsc{mk - bg}}\vspace{-0.1cm}
    \end{subfigure}
    \hfill
    \begin{subfigure}[t]{0.27\textwidth}
        \raisebox{-\height}{\includegraphics[width=\textwidth]{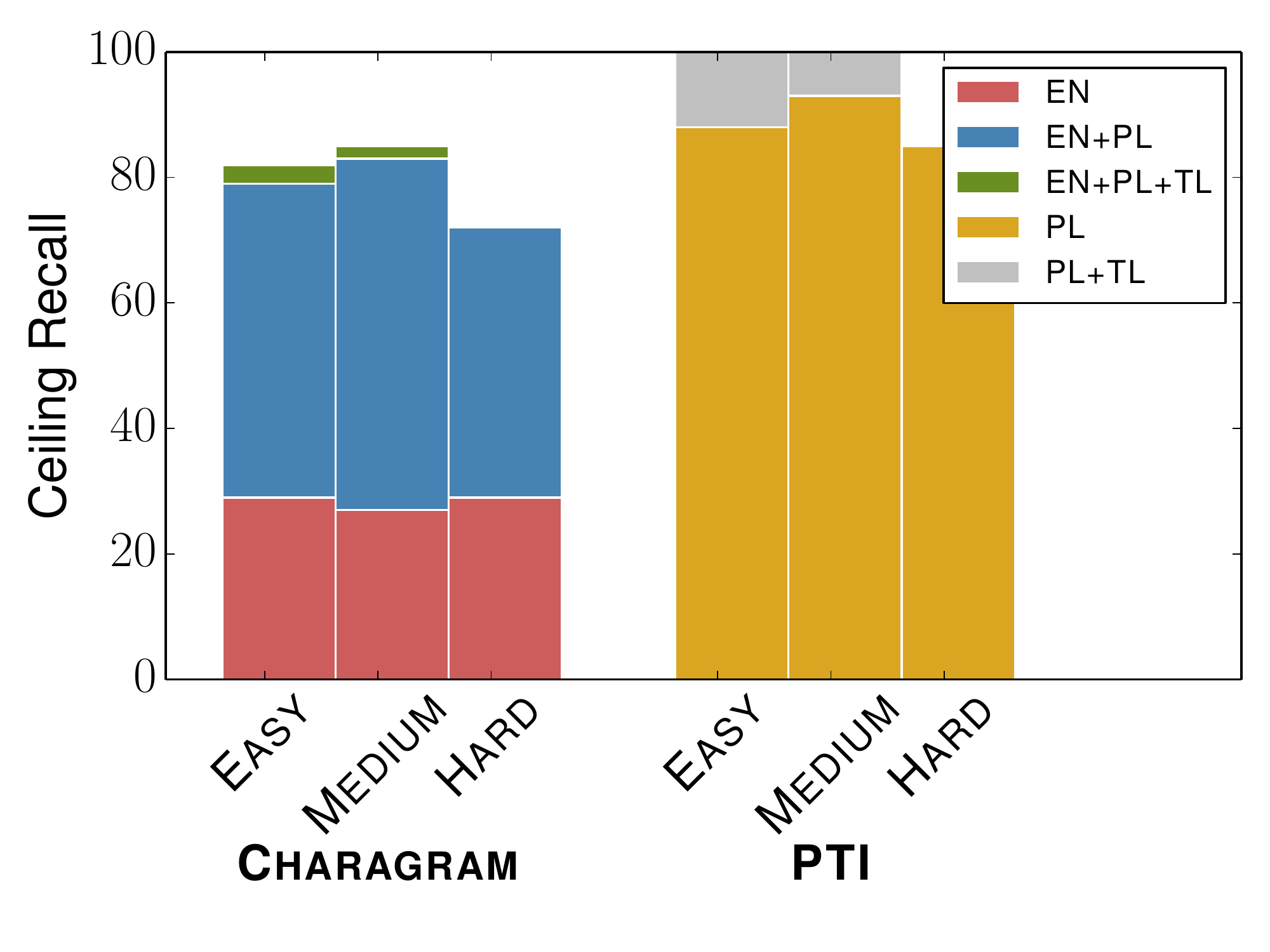}}
        \vspace{-0.4cm}\caption{\textsc{jv - id}}\vspace{-0.1cm}
    \end{subfigure}
    \moveups
    \begin{subfigure}[t]{0.27\textwidth}
        \raisebox{-\height}{\includegraphics[width=\textwidth]{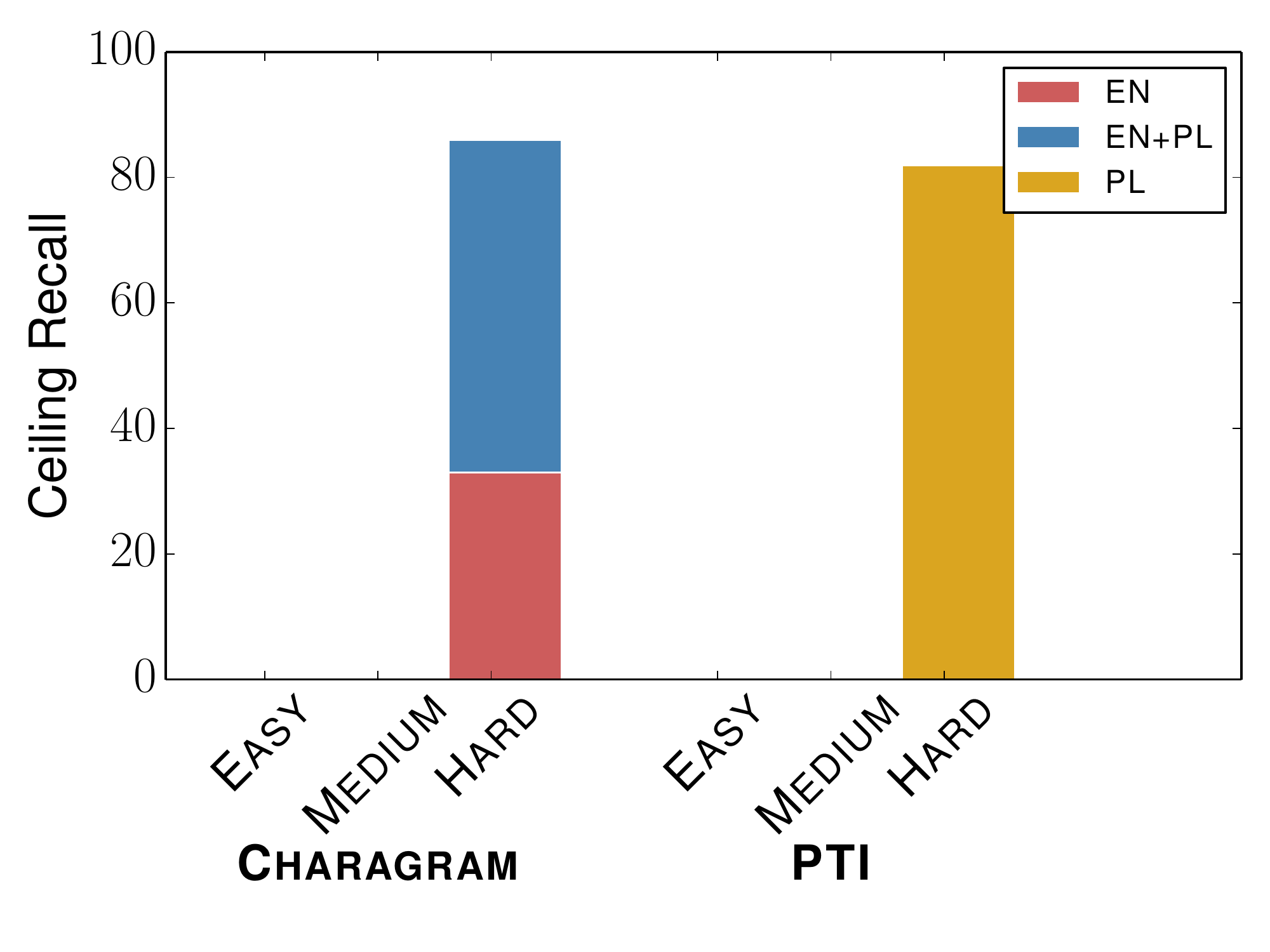}}
        \vspace{-0.4cm}\caption{\textsc{tk - az}}
    \end{subfigure}
    \hfill
    \begin{subfigure}[t]{0.27\textwidth}
        \raisebox{-\height}{\includegraphics[width=\textwidth]{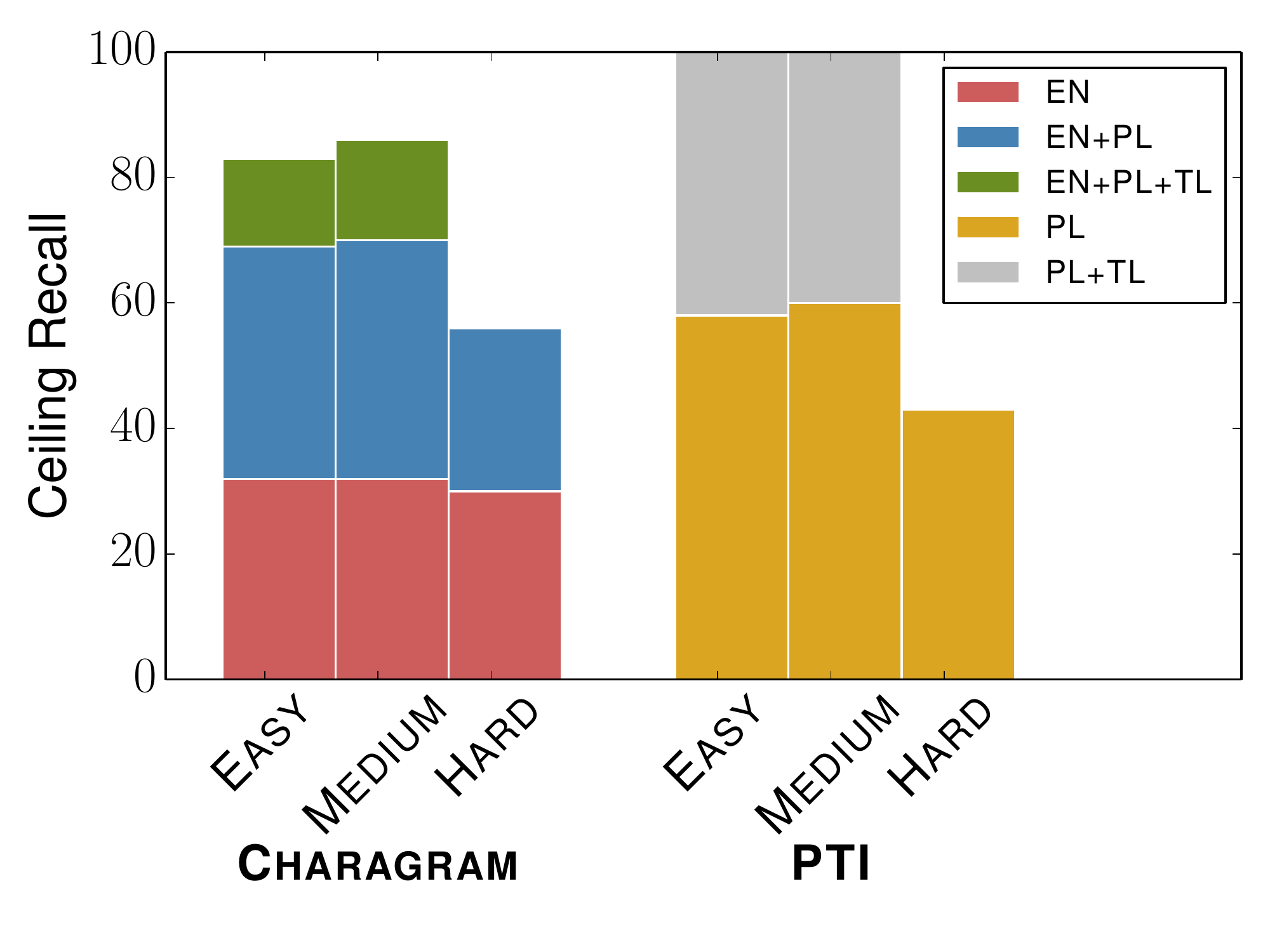}}
        \vspace{-0.4cm}\caption{\label{fig:weak1}\textsc{mr - hi}}
    \end{subfigure}
    \hfill
    \begin{subfigure}[t]{0.27\textwidth}
        \raisebox{-\height}{\includegraphics[width=\textwidth]{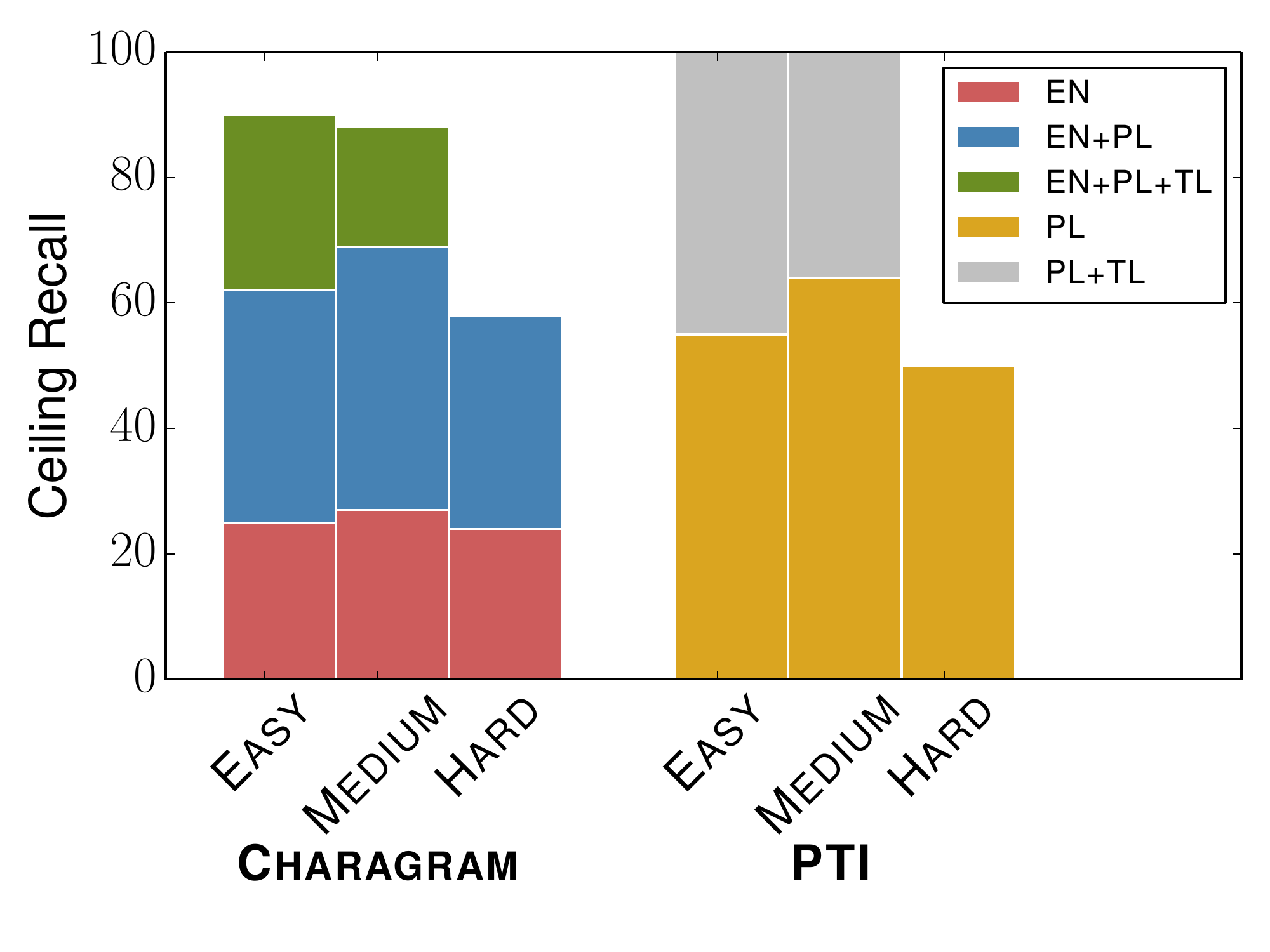}}
        \vspace{-0.4cm}\caption{\label{fig:weak2}\textsc{fo - is}}
    \end{subfigure}
    \figcaption{\label{ceilings} Ceiling recall for all query types and target languages. In the experiments reported in Section \ref{sec:exps}, the candidate space of \textsc{pti} is as follows: \textsc{PL} and \textsc{PL+TL} in the zero-shot and supervised setting, respectively. Whereas for \textsc{Charagram} is as follows: \textsc{EN+PL} and \textsc{EN+PL+TL} in the zero-shot and supervised setting, respectively.} 
\end{figure*}

For some methods such as \textsc{Charagram} is easy to expand the candidate space by considering entities in Wikipedias other than English. While one might evaluate all entities (more than 20 million), we discovered one much less costly alternative. We experimentally observe that expanding the candidate space with the entities observed in the training data of the pivot language significantly improves the entity coverage of the entities observed in the test data of the target language (more in Section \ref{sec:analysis}). With the goal of performing a realistic evaluation we simply evaluate all mentions.


\subsection{Training Details}
\label{sec:training}

The simplicity of our approach \textsc{pti} is also reflected in its number of hyperparameters. The hyperparameter $\alpha$, which controls the contribution of the pivot language in the construction of the prior and posterior probabilities, is validated among the values $\{0.1, 0.4, 1\}$. The hyperparameter $\lambda$, which controls the importance of the posterior probability, is validated among the values $\{0.2, 0.4, 1\}$. As in \textsc{Charagram}, the tokenizer \texttt{tknr} applied to mentions returns character $n$-grams with $n \in \{2, 3, 4, 5\}$. The tokenizer counts each symbol as a character regardless of the script of the language, although more sophisticated mechanisms are also possible (\eg by converting strings to international phonetic alphabet (IPA) symbols). For \textsc{Charagram} we replicate the exact same setup as reported in \cite{Zhou2020ImprovingCG}. The hyperparamenter $\mu$ of the supervised setting is validated among the values $\{0.01, 0.1, 0.2\}$. More details are provided in Appendix \ref{app:wikipriors}. The hyperparameters $\alpha$ and $\mu$ are required by both \textsc{pti} and \textsc{Charagram}, respectively, only in the supervised setting. For all cases, the validation metric is micro recall@30. 


\section{Results}
\label{sec:results}
Table \ref{table:main_results} depicts a comparison of the methods in terms of micro recall@30. Similar conclusions are drawn for both learning settings. Overall, our approach \textsc{pti} clearly outperforms \textsc{Charagram} in those target languages where the pivot language is high-resourced. For pivot languages that are medium-resourced, while \textsc{pti} is the best performing technique \textsc{Charagram} performs very similarly in one target language (\textsc{mk}). This tendency is not as clear when the pivot language is low-resourced, where \textsc{Charagram} shows a clear superior performance in the target language \textsc{mr} in the zero-shot learning setting. We see that all models largely benefit from the existence of training data in the target language. 

By the definition of the query types introduced in Section \ref{sec:query}, it follows that all queries are of type \textsc{hard} in the zero-shot setting. However, this is not the case when the models are also trained with data from the target language. We show the performance breakdown in Table \ref{table:results_types}. The conclusions are clear: \textsc{pti} always outperforms \textsc{Charagram} for queries of type \textsc{Easy} and \textsc{Medium}, and it is only beaten for queries of type \textsc{Hard} when the pivot language has low resources, with the exception of \textsc{az}. We note that while in this work we have created the evaluation splits so that all query types are equally important, the distribution of the query types depends on the target evaluation data. For instance, splitting Wikipedia articles into training, validation and test in a proportion of 70\%, 15\% and 15\%, respectively, leads to a distribution of query types largely dominated by \textsc{Easy} ones.

The performance of \textsc{WikiPriors} is also dependent on the amount of resources of the pivot language, showing a performance that is comparable or even better than \textsc{Charagram} when the pivot language is high- and medium-resourced (see Table \ref{table:main_results}). The more detailed analysis depicted in Table \ref{table:results_types} shows that \textsc{WikiPriors} has a very heterogeneous performance with respect to the different query types in the supervised setting. As expected, it obtains a perfect performance for \textsc{Easy} queries, but its performance tends to be poor, with some few exceptions, for \textsc{Medium} and \textsc{Hard} queries. 


\subsection{Analysis}
\label{sec:analysis}

\noindent{\textbf{Entity Coverage}}
As argued in Section \ref{sec:realistic}, it is unrealistic to simply ignore those mentions that are not in the considered KB, specially if this translates into a performance that would not resemble that of the realistic evaluation that we propose. We proceed to analyze the ceiling recall---which depends on the entity coverage---of the methods in the realistic setting, where no mention is excluded.

\begin{figure}[t]
     \centering
    \begin{subfigure}[t]{0.23\textwidth}
        \raisebox{-\height}{\includegraphics[width=\textwidth]{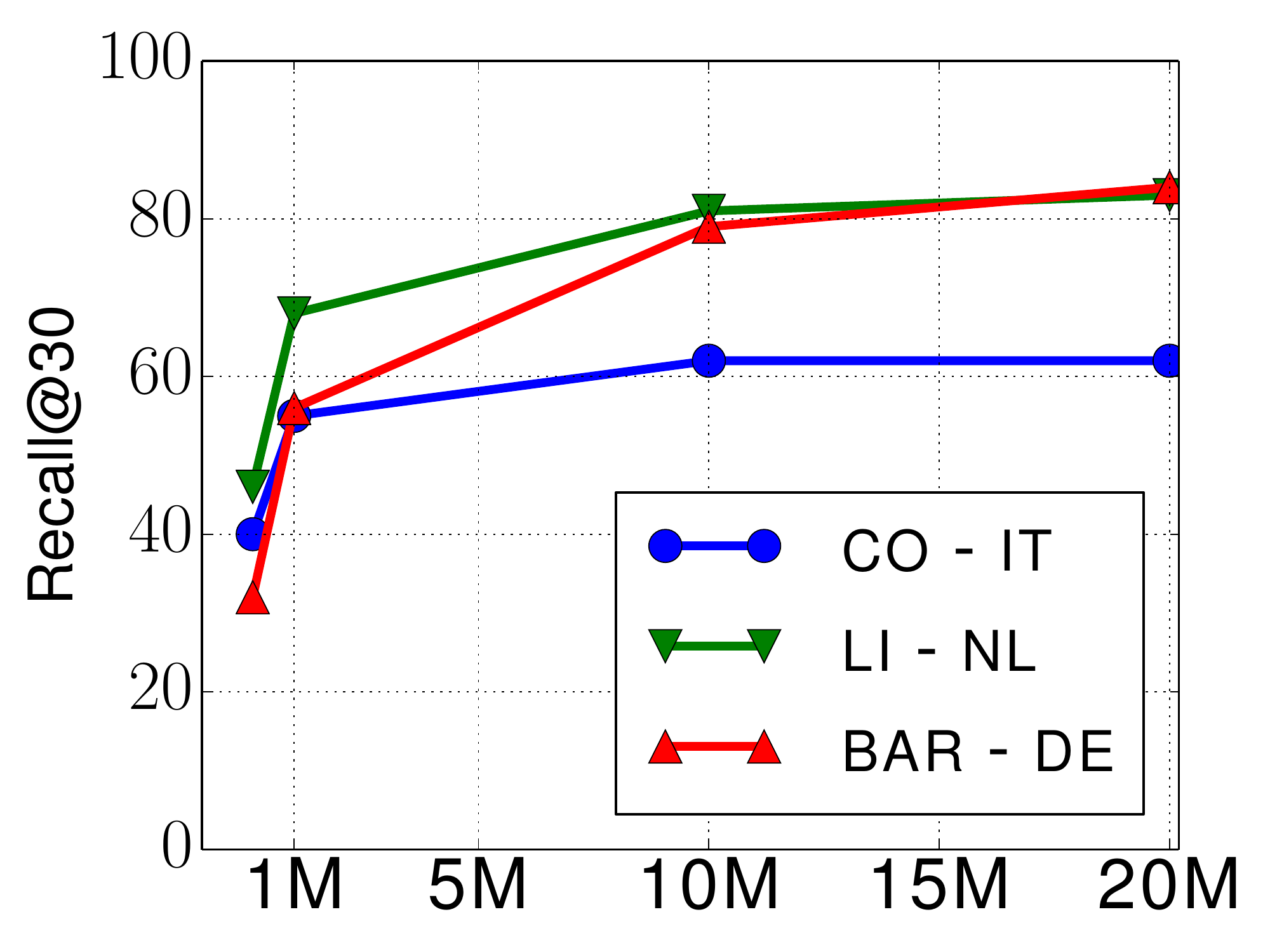}}
        \caption{\textsc{pti}}
    \end{subfigure}
    \hfill
    \begin{subfigure}[t]{0.23\textwidth}
        \raisebox{-\height}{\includegraphics[width=\textwidth]{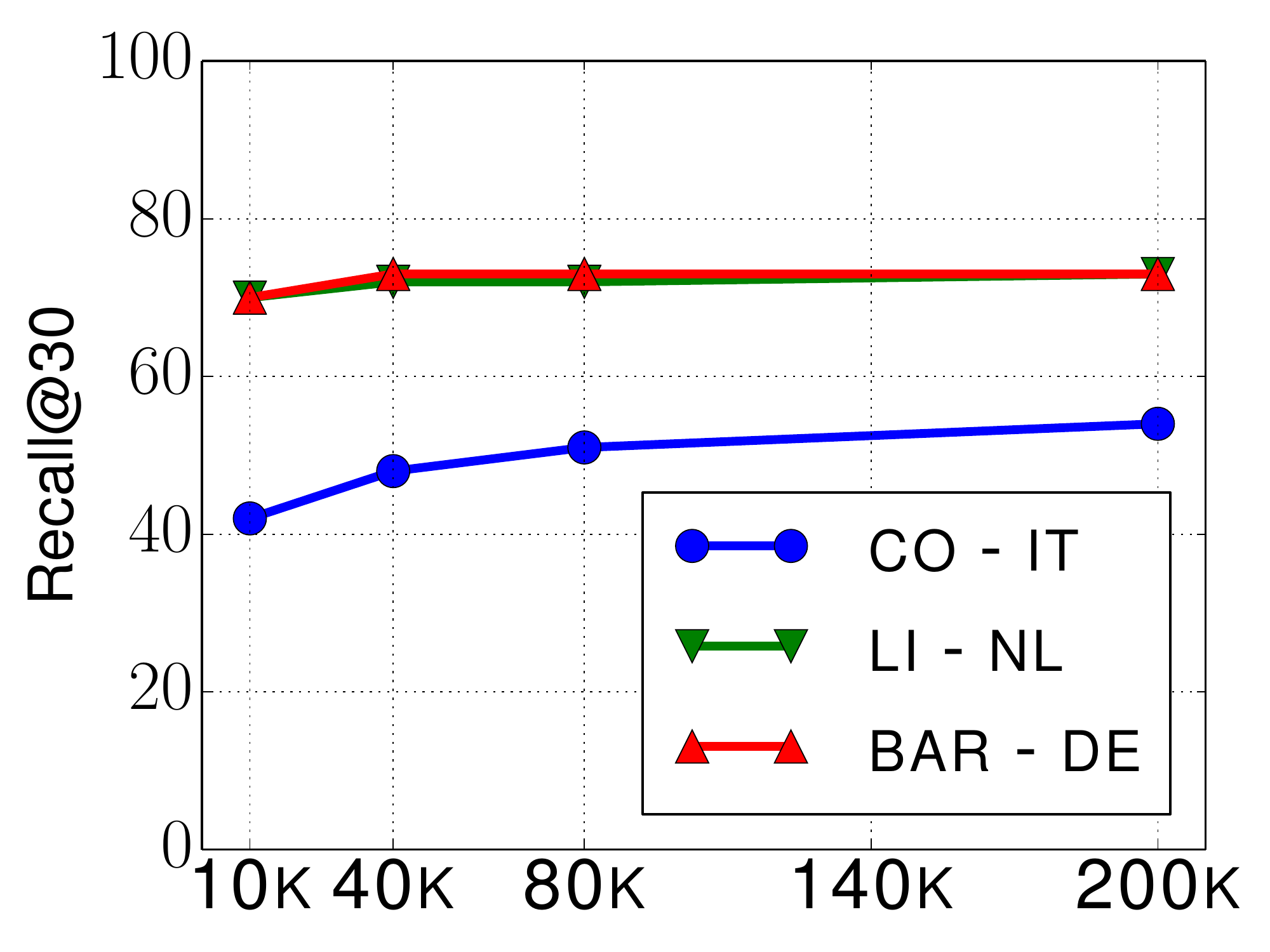}}
        \caption{\textsc{Charagram}}
    \end{subfigure}
    \figcaption{\label{perf_data} Recall@30 of \textsc{pti} (left) and \textsc{Charagram} (right) for different amount of training data. For simplicity, the candidate space of \textsc{Charagram} is always formed as if all training data was observed.}
\end{figure}

Figure \ref{ceilings} shows the ceiling recall of \textsc{pti} and \textsc{Charagram} for the different target languages and query types. The ceiling recall of \textsc{Charagram} is lower than 30\% in almost all cases when the candidate space is given by the English Wikipedia (\textsc{EN}) considered in previous work \cite{Rijhwani2019ZeroshotNT,Zhou2020ImprovingCG}. This showcases that previous work were limiting, to a great extent, the mention queries that were evaluated. This is not only unrealistic, but also a notable limiting factor in performance. The ceiling recall notably increases when the candidate space also includes the entities that are observed in the training data of the pivot language. We hypothesize that this is because the pivot language is not only lexically similar to the target language, but also their respective Wikipedias exhibit similar topical interests. Moreover, the ceiling for queries of type \textsc{Easy} and \textsc{Medium} is also increased when the candidate space is also formed by the entities observed in the training data of the target language. The latter corresponds to the supervised setting; and surprisingly, as opposed to \textsc{pti}, \textsc{Charagram} does not reach a ceiling recall of 1 for queries of type \textsc{Easy} and \textsc{Medium}. The reason is that there are a fair amount of entities in the respective Wikipedias that do not have an English name (\eg \textsc{Q61068957} or \textsc{Q15842755}). This is the strength and weakness of \textsc{Charagram}: its candidate space can include any entity as long as its English title is available. On the other side, \textsc{pti} is language-agnostic to the representation of the entity, but its entity coverage is limited to the entities observed in the pivot language (\textsc{PL}), or in both the pivot and target language (\textsc{PL+TL}). It is for this reason that \textsc{pti} is beaten in queries of type \textsc{Hard} when the pivot language is low-resourced : it fully relies on the entity coverage of the pivot language, which is low, to successfully answer these queries. This is illustrated in Figure \ref{fig:weak1} and \ref{fig:weak2}, where \textsc{Charagram} shows a higher ceiling recall for \textsc{Hard} mentions due to the additional entity coverage provided by the English (\textsc{EN}) Wikipedia. Future work should address this limitation of \textsc{pti} (\eg by incorporating more pivot languages). However, in all other situations where the ceiling recall is higher or comparable to that of \textsc{Charagram}, \textsc{pti} shows the best performance. 

\noindent{\textbf{Impact of the amount of training data}} We train both \textsc{pti} and \textsc{Charagram} with increasing amount of training data, and computed micro recall@30 for several target languages. This is shown in Figure \ref{perf_data}. The performance of \textsc{pti} increases with the more data up to a point---around 10 million. One reason is that the entity coverage of \textsc{pti} is expected to increase with the amount of training data. Another reason is that the constructed indexes become more accurate. On the other side, \textsc{Charagram} barely improves beyond 80,000 data points\footnote{Authors of \textsc{Charagram} confirmed in an email they used, at most, this amount of data for training}. \textsc{Charagram} needs relatively little data to learn equivalences between $n$-grams co-occurring in mentions and English entity names, but gets stuck very quickly and does not leverage all the statistical information contained in the training corpora.  

\noindent{\textbf{Runtime and Memory}} The goal of a candidate generator is to reduce the computational cost of a more complex subsequent model. Therefore, besides recall, complexity must also be a factor that drives the design of a candidate generator. We perform a detailed analysis of the complexity in Appendix \ref{app:complexity}. The conclusion of such analysis is that in average, \textsc{pti} requires 20 times less memory than \textsc{Charagram}, and it is 300 and 7 times faster than \textsc{Charagram} in terms of training and inference time, respectively.

\begin{table}[t]
\centering
\vspace{1mm}
\tabcaption{\label{table:results_high} Recall@30 in the high-resource setting. }
\resizebox{0.8\linewidth}{!}{
    \begin{tabular}{lrrrrrr}
    \toprule
     \textbf{Lang.} & \multicolumn{2}{c}{\textsc{it}} & \multicolumn{2}{c}{\textsc{nl}}  &    \multicolumn{2}{c}{\textsc{de}} \\ 
     \cmidrule(lr){2-3} \cmidrule(lr){4-5} \cmidrule(lr){6-7} 
     \textbf{Query Type} & \textsc{E} & \textsc{M} & \textsc{E} & \textsc{M} & \textsc{E} & \textsc{M} \\ 
          \midrule    
    \textsc{WikiPriors} & \textbf{95} & 32 & \textbf{98} & 30 & \textbf{98} & 34 \\
     \textsc{Charagram} & 50 & 45 & 57 & 47 & 39 & 38 \\ \cmidrule(lr){1-7}
     \textsc{pti} & 79 & \textbf{49} & 87 & \textbf{53} & 77 & \textbf{54} \\
        \bottomrule
    \end{tabular}
    }
    \moveup
\end{table}


\subsection{Can \textsc{pti} also become the \textit{de-facto} choice for high-resource languages?}
\label{sec:high}
We analyze whether \textsc{pti} may compete with \textsc{WikiPriors} in the most standard candidate generation setting: candidates have to be generated for mentions of a high-resource language by only exploiting its own training corpus. We build this experiment motivated by the recent findings disclosed in the recent work by \textit{Fu et al.} \cite{Fu2020DesignCF}, where authors found that even high-resources languages, such as Spanish, are also limited by their mention coverage. This limitation relates to the queries of type \textsc{Medium}.  \textit{Fu et al.} partially circumvent the limitation with information coming from search engine query logs.

 Following the same protocol as in Section \ref{sec:data} we create 1,000 queries of type \textsc{Easy} and \textsc{Medium} for the three high-resource languages used in Section \ref{sec:exps} as pivot languages: Italian (\textsc{it}), Dutch (\textsc{nl}) and German (\textsc{de}). We do not include queries of type \textsc{Hard} as they cannot be successfully completed. The results, depicted in Table \ref{table:results_high}, indicate that \textsc{pti} always obtains the best performance for queries of type \textsc{Medium}, followed by \textsc{Charagram}. On the other side, for queries of type \textsc{Easy} standard lookup tables (\ie \textsc{WikiPriors}) exhibit the best performance. We conclude that for the standard setting, \textsc{WikiPriors} must be used as the primary technique, falling back to \textsc{pti} only if \textsc{WikiPriors} fails to retrieve any candidate.

\section{Conclusions}
We perform an in-depth analysis of the candidate generation problem. We show that the inherited practice of ignoring mentions where the target entity is not in the English Wikipedia is unrealistic in a cross-lingual setting. We alleviate this problem with an efficient solution that consists of using the pivot and target language to expand the candidate space. We also contribute with a categorization of the mention queries according to their difficulty. Finally, we propose a \textit{light-weight} approach, called \textsc{pti}, that outperforms the current state-of-the-art in almost all target languages and query types.

\section*{Acknowledgements}
This project was partly funded by the Swiss National Science Foundation (grant 200021\_185043), the European Union (TAILOR, grant 952215), and the Microsoft Swiss Joint Research Center. We also gratefully acknowledge generous gifts from Facebook and Google supporting West’s lab.

\section{Bibliographical References}\label{reference}

\bibliographystyle{lrec2022-bib}
\balance
\bibliography{main}


\clearpage
\appendix

\section{Supplementary Material}
\label{sec:appendix}

\subsection{Dataset Information}
\label{app:dataset}
We preprocess the Wikipedia dumps for each of the target and pivot languages. We apply possible redirects to each Wikipedia entity, and map the (redirected) Wikipedia entity to the corresponding Wikidata entity\footnote{This last step is not really necessary, but we prefer a language-agnostic representation of the entities.} (\eg \textsc{Q21} is the Wikidata identifier for the Wikipedia entity \textsc{England}).  We do not use aliases \cite{Zhou2020ImprovingCG} to enrich the data. Nevertheless its inclusion will improve the performance of \textit{any} technique. 

\begin{table}[!htb]
\centering
\normalsize
\vspace{0.1cm}
\tabcaption{\label{table:data} Pivot languages used in this work categorized according to their amount of resources: high-, medium- and low-resource are in the upper, medium and lower block, respectively.}
\resizebox{0.8\columnwidth}{!}{
    \begin{tabular}{lrr}
    \toprule
     \textbf{Lang. } & \textbf{WP Size} & \textbf{\#Mention-Entity pairs.}  \\ \midrule  
     \textsc{it} (Italian) & 1.5M & 24M \\
     \textsc{nl} (Dutch) & 2M & 16M \\
     \textsc{de} (German) & 2.5M & 44M \\ \hdashline
     \textsc{fi}  (Finnish) & 500k & 5.7M \\
     \textsc{bg}  (Bulgarian) & 250k & 3.4M \\
     \textsc{id} (Indonesian) & 500k &  4.4M \\ \hdashline
     \textsc{az}   (Azerbaijani)& 150k & 1.1M \\
     \textsc{hi}  (Hindi) & 140k & 700k \\
     \textsc{is}  (Icelandic) & 50k & 400k \\ 
        \bottomrule
    \end{tabular}}
\end{table}

The transfer learning paradigm of applying a model, learned in a pivot language, to a target language has been explored in different natural language problems. Nevertheless, such works have never reached a consensus on how to select a pivot language that is closely related to a target language. For those target languages that have been used in previous work (\eg \textsc{jv} or \textsc{mr}) we select the same pivot languages as in these works. For all others we rely on \textsc{lang2vec} \cite{Levin2017URIELAL} to choose appropriate pivot languages. 

\subsection{More Details about \textsc{WikiPriors} and \textsc{Charagram}}
\label{app:wikipriors}
\textsc{WikiPriors} is based on the candidate generator used in previous work \cite{Tsai2016CrosslingualWU,Upadhyay2018JointMS}. It fallbacks to the pivot language when the index of the target language does not retrieve $K$ candidates. On the contrary, for the zero-shot setting it only relies on the index constructed using the pivot language. Furthermore, for both zero-shot and supervised settings, it constructs indexes that associate the constituent words of mention strings with entities using prior probabilities $P(\text{entity}|\text{word}), \forall word \in mention$. These indexes are used to retrieve candidates if the indexes on mention strings provide less than $K$ candidates. Previous work \cite{Zhou2020ImprovingCG} has shown that sometimes it is competitive as compared to \textsc{Charagram}.\\

We replicate the exact same setup for \textsc{Charagram} as reported in \cite{Zhou2020ImprovingCG}: the tokenizer \texttt{tknr}, which is applied to mentions and (English) entity names, returns character $n$-grams with $n \in \{2, 3, 4, 5\}$. The embedding size is 300. We train the model with stochastic gradient
descent (SGD) with batch size 64, and a learning rate of 0.1. We stop training if the micro recall@30 on the validation set does not increase for 50 epochs, and the maximum number of training epochs is set to 200.

\begin{table}[h]
\centering
\caption{\label{table:thresholding} Micro recall@30 after applying thresholding. In parenthesis the percentage of entries that are removed from the indexes built by \textsc{pti}.}
\small
\resizebox{0.8\columnwidth}{!}{
    \begin{tabular}{lrrr}
    \toprule
            & \multicolumn{3}{c}{\textbf{Thresholding Value}} \\ \cmidrule(lr){2-4}
     \textbf{Target - Pivot} & 0 & 1e-2 & 0.1 \\ \midrule  
     \textsc{co} - \textsc{it} & 63 (0\%)& 62 (72\%) & 48 (98\%) \\
     \textsc{li} - \textsc{nl}  & 84 (0\%) & 83 (65\%) & 73 (98\%) \\
     \textsc{bar} - \textsc{de} & 88 (0\%) & 87 (75\%) & 70 (99\%) \\
     \textsc{olo} - \textsc{fi}   & 72 (0\%) & 71 (68\%) & 55 (97\%) \\
     \textsc{mk} - \textsc{bg}  & 71 (0\%) & 69 (66\%) & 55 (95\%) \\
     \textsc{jv} - \textsc{id} & 87 (0\%) &  87 (64\%) & 79 (96\%) \\
     \textsc{tk} - \textsc{az}   & 69 (0\%) & 69 (56\%) & 64 (92\%) \\
     \textsc{mr} - \textsc{hi}  & 70 (\%0) & 69 (62\%) & 58 (94\%) \\
     \textsc{fo} - \textsc{is}  & 79 (\%0) & 79 (53\%) & 73 (90\%) \\ 
        \bottomrule
    \end{tabular}}
\end{table}

\subsection{Complexity of \textsc{pti} and \textsc{Charagram}}
\label{app:complexity}
All the experiments were done using code written in Python on an Intel(R) Xeon(R) E5-2680 24-core machine with 2.50GHz CPU, and 256 GB RAM running Linux Ubuntu 16.04. Experiments for \textsc{Charagram} are run in a TitanX GPU. We do not make use of any parallelism in the experiments.

As previously discussed, the candidate generator must be computational efficient. For this reason we compare the memory and time complexity of \textsc{pti} to that of \textsc{Charagram}. We experimentally observe that it is possible to reduce significantly the amount of non-zero prior and posterior probabilities in \textsc{pti} by thresholding (see Table \ref{table:thresholding}), while having very little impact on the performance. These experiments correspond to the best performing learning setting. We perform further analysis of \textsc{pti} with a thresholding value of 0.01, and compare it to \textsc{Charagram} (trained with 80,000 data points) in terms of number of parameters, training and inference time. The inference time corresponds to the time needed to evaluate all mention queries---3,000 queries for almost all target languages. All target languages are evaluated in the best performing learning setting. A comparison is shown in Table \ref{table:complexity}. The findings are summarized below:

\begin{figure*}[!htb]
\begin{minipage}{0.6\linewidth}
\resizebox{1\columnwidth}{!}{
    \begin{tabular}{lrrrrrr}
    \toprule
    & \multicolumn{2}{c}{\textbf{Memory}} & \multicolumn{4}{c}{\textbf{Time}} \\ \cmidrule(lr){2-3} \cmidrule(lr){4-7}
            & \multicolumn{2}{c}{\textbf{\#Parameters}}
            & \multicolumn{2}{c}{\textbf{Training} (minutes)} & \multicolumn{2}{c}{\textbf{Inference} (seconds)} \\ \cmidrule(lr){2-3} \cmidrule(lr){4-5} \cmidrule(lr){6-7}
     \textbf{Target - Pivot} & \textsc{pti} & \textsc{Charagram} & \textsc{pti} & \textsc{Charagram} & \textsc{pti} & \textsc{Charagram}\\ \midrule  
     \textsc{co} - \textsc{it} & 45M & 200M & 10 & 180 & 45 & 85 \\
     \textsc{li} - \textsc{nl}  & 38M & 220M & 7 & 495 & 60 & 120 \\
     \textsc{bar} - \textsc{de} & 80M & 225M & 20 & 645 & 150 & 150 \\
     \textsc{olo} - \textsc{fi} & 18M & 180M & 3 & 420 & 12 & 70 \\
     \textsc{mk} - \textsc{bg}  & 10M & 225M & 1 & 770 & 30 & 120 \\
     \textsc{jv} - \textsc{id} & 11M & 200M & 2 & 480 & 20 & 120 \\
     \textsc{tk} - \textsc{az} & 4M & 160M & $<$1 & 250 & 3 & 60 \\
     \textsc{mr} - \textsc{hi}  & 3.5M & 150M & $<$1 & 680 & 10 & 110 \\
     \textsc{fo} - \textsc{is}  & 2.2M & 120M & $<$1 & 480 & 6 & 115\\ 
        \bottomrule
    \end{tabular}}
\captionof{table}{\label{table:complexity} Comparison between \textsc{pti} (with a thresholding value of 0.01) and \textsc{Charagram} (trained with 80,000 data points as in \protect\cite{Zhou2020ImprovingCG} in terms of number of parameters, training and inference time.}
\end{minipage}
\hfill
\begin{minipage}{0.35\linewidth}
\includegraphics[width=\columnwidth]{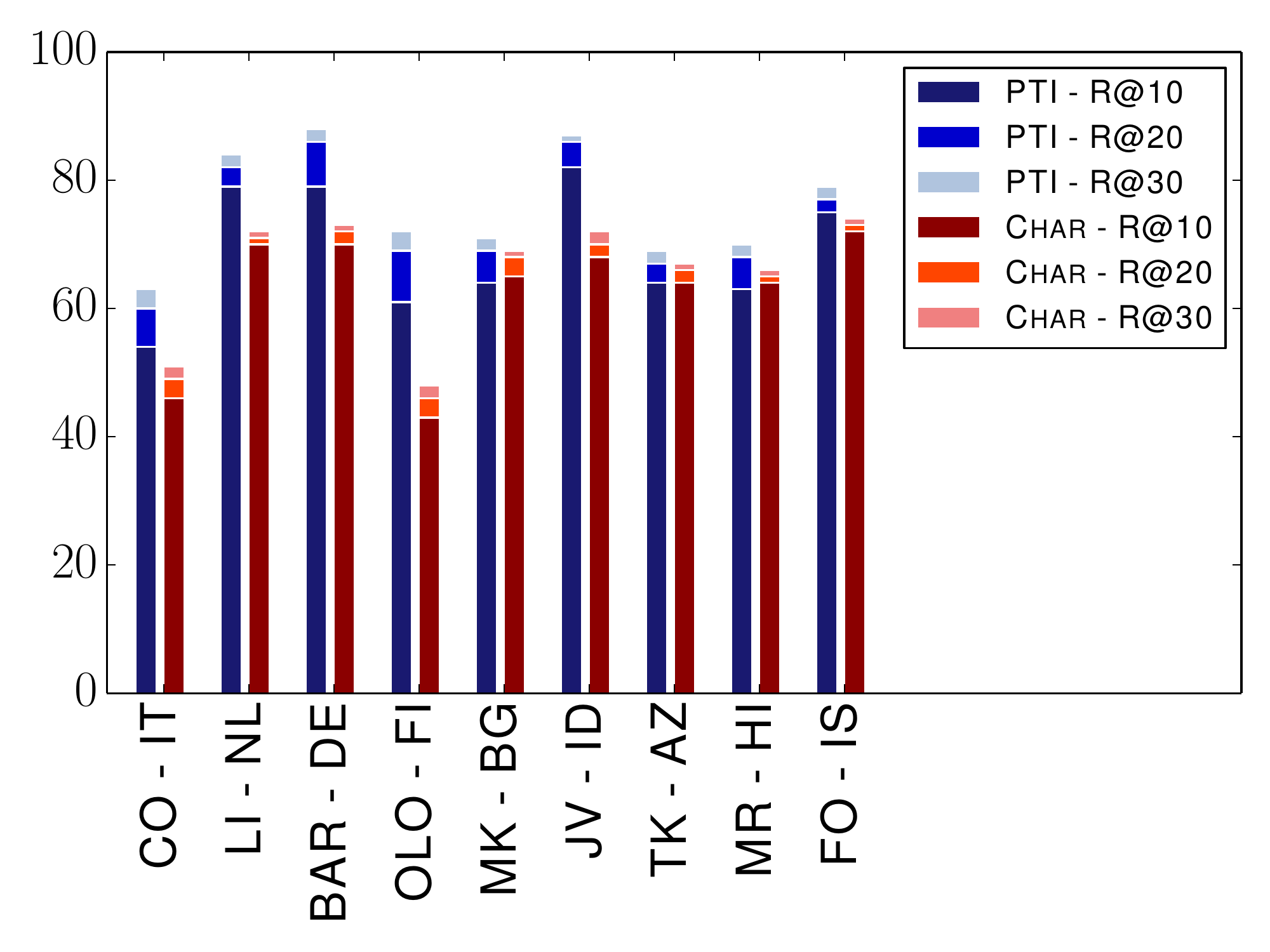}
    \caption{\label{fig:recall} Recall@$K$ (\textsc{R@}$K$) of \textsc{pti} and \textsc{Charagram} (\textsc{Char}) for several values of $K$.}
\end{minipage}
\end{figure*}

\begin{itemize}[nolistsep,leftmargin=4mm]
    \item \textbf{Number of parameters}. While the number of parameters remains more or less constant for \textsc{Charagram}, \textsc{pti} shows its number of parameters is proportional to the amount of resources of the pivot language. 
    \item \textbf{Training time}. The training time of \textsc{pti} corresponds to the time that takes to construct the prior and posterior probabilities, which is proportional to the amount of resources of the pivot language. The optimization of the learning objective done by \textsc{Charagram} is significantly much slower.
    \item \textbf{Inference time}. \textsc{pti} uses sparse matrices to represent $P(\text{entity}|\text{token})$ and $P(\text{token}|\text{entity})$. At test time, scores are simply computed by aggregating rows of such matrices. Moreover, the search performed by \textsc{pti} to find the $K$ largest values only includes those candidates with non-zero scores. These characteristics make often \textsc{pti} more efficient than \textsc{Charagram} during inference.
\end{itemize}

\subsection{Recall@$K$ for different values of $K$}
\label{app:recall}
Previous works on candidate generation have set a value of 30 for $K$. However, when the candidate generator is followed by an entity disambiguation technique, there are other values that are typically used by these methods such as (approximately) 10 \cite{Ganea2017DeepJE} and 20 \cite{Upadhyay2018JointMS}. Figure \ref{fig:recall} shows the recall of \textsc{pti} and \textsc{Charagram} for these other values. Except for \textsc{CO-IT}, \textsc{OLO-FI} and \textsc{TK-AZ}, where zero-shot is the only possible learning setting, all other metrics are evaluated in the supervised setting. The comparative performance between the methods is similar for all values of $K$.


\subsection{Other versions of \textsc{pti}}
\label{app:failed_ext}
\noindent\textbf{Wildcard Tokens} The tokenizer extends the tokens returned for a given mention with wildcard tokens. Wildcard tokens includes placeholders represented by an asterisk, which can be interpreted as any character. We observed some minor improvements in recall for some target languages, but this notably increased the amount of tokens and, consequently, the memory requirements of the approach.

\noindent\textbf{Probability-level Fusion} For the supervised learning setting, we explored another approach to integrate the information from both the target and pivot language. Independent prior and posterior probabilities are computed for both languages, and linearly combined with a weighting factor. Finally, the probabilities are normalized. We did not observe any improvement in performance with this fusion technique.

\noindent\textbf{Probability Smoothing} We perform one additional step in the previous version. We use additive smoothing \cite{ngo2004smoothing} (also known as Laplace smoothing) in the computation of the prior and posterior probabilities for the pivot language. These smoothed probabilities account for possible discrepancies between the target and pivot language. We sometimes observed a minor improvement for some query types, but this comes at the cost of validating one more hyperparameter (the smoothing factor).

\end{document}